\documentclass[a4paper]{article}
\usepackage{cite}
\usepackage{amsmath, amssymb, amsfonts}
\usepackage{algorithm}
\usepackage{algorithmic}
\usepackage{mdwlist, multicol}
\usepackage{array, booktabs, enumitem, multirow}
\usepackage{graphicx}
\usepackage{float}
\usepackage{textcomp}
\usepackage{xcolor}
\usepackage{subfigure}
\usepackage{epsfig}
\usepackage{INTERSPEECH2021}
\title{Federated Learning with Dynamic Transformer for Text to Speech} 
\name{Zhenhou Hong, Jianzong Wang*\thanks{* Corresponding author: Jianzong Wang, jzwang@188.com.}, Xiaoyang Qu, Jie Liu, Chendong Zhao, Jing Xiao}
\address{Ping An Technology (Shenzhen) Co., Ltd.}
\email{}

\begin{document}

\maketitle

\begin{abstract}
Text to speech (TTS) is a crucial task for user interaction, but TTS model training relies on a sizable set of high-quality original datasets. 
Due to privacy and security issues, the original datasets are usually unavailable directly. 
Recently, federated learning proposes a popular distributed machine learning paradigm with an enhanced privacy protection mechanism. 
It offers a practical and secure framework for data owners to collaborate with others, thus obtaining a better global model trained on the larger dataset. 
However, due to the high complexity of transformer models, the convergence process becomes slow and unstable in the federated learning setting. 
Besides, the transformer model trained in federated learning is costly communication and limited computational speed on clients, impeding its popularity.
To deal with these challenges, we propose the federated dynamic transformer. 
On the one hand, the performance is greatly improved comparing with the federated transformer, approaching centralize-trained Transformer-TTS when increasing clients number. 
On the other hand, it achieves faster and more stable convergence in the training phase and significantly reduces communication time. 
Experiments on the LJSpeech dataset also strongly prove our method's advantage.

\end{abstract}

\section{Introduction}
\label{sec:intro}
Text to speech (TTS)~\cite{zeng2020aligntts,sun2020graphtts,sun2021graphpb} synthesis is an active research area. 
Despite decades of research, generating natural speech from text is still a challenging task. With the development of voice technology in recent years, various voice services and applications have new advances, affecting people's daily lives. 
Over time, different technologies have taken over the field. 
There are now feasible ways to produce natural prosody with high audio fidelity using a much-simplified voice building pipeline~\cite{atten,embed,intera,bert}. 
Recently, many variants of transformer models such as FastSpeech~\cite{ren2019fastspeech}, Transformer-TTS~\cite{li2019neural}, AlignTTS~\cite{zeng2020aligntts}, MultiSpeech~\cite{multi} have shown great promises in the TTS task.  
However, such transformer models typically require sizable high-quality training data to achieve a good performance.
Nevertheless, limitations like data privacy, liability, and regulatory concerns may make it difficult for clients to collaborate with other data owners for centralized aggregation training. 

To tackle this problem, federated learning~\cite{mcmahan2017communication,kairouz2019advances,zhao2021efficient,liu2021quantitative} has presented as a paradigm that allows clients to train a shared global model collaboratively without data breaches. 
When training models in the federated learning setting, participating clients do not upload their local data to the central server; instead, a central aggregator coordinates the optimization procedure among the clients~\cite{fed,fede}. 
At each iteration of this procedure, clients compute gradient-based optimization using their local data to update the current model parameters and then upload only these model's updates to the central aggregator. 
The federated learning framework promises to enhance the model's performance by multipartite data aggregation under clients' absolute privacy conditions.

Federated learning has proved to be a compelling framework in different areas, such as optical character recognition (OCR)~\cite{zhu2019federated}, natural language processing (NLP)~\cite{DBLP:conf/aaai/LiuWGFZ20,zhu2020empirical}, Medical Imaging~\cite{kaissis2020secure} and so on. 
These show that federated learning has a promising prospect in further practical applications. 
However, little researches have been done in the TTS area, and especially the model is as large as a transformer. 
To be specific, the main impediments are as follows: Firstly, all distributed learning methods fatally suffer from the slow and easily unstable convergence problem, which is mainly caused by different clients' non-IID(including quantity skew)~\cite{kairouz2019advances,noniid}. 
When larger amounts of parameters are contained, the model will be more challenging to converge. 
Secondly, training such a large model is costly in the federated learning setting. 
Large amounts of parameters aggravate the computational burden locally and extremely increase the communication time and cost. 

To address these issues, we propose the Dynamic Transformer (FedDT-TTS) with faster convergence speed and lower communication cost in the federated learning framework for the TTS task. 
The key insight is, by changing the layer-wise training in the wake-sleep algorithm~\cite{hinton2006fast}, we grow both the encoder and decoder dynamically through the training process. 
This dynamic adding layers mechanism makes the shallow layers in encoder and decoder faster of learning low-level texts or mel features, thus allowing the deeper layers in encoder and decoder to learn high-level text or mel feature information more easily. 
Moreover, in order to stabilize the convergence, we let the new layer initialize with weights divided by a per-layer normalize constant according to He's initializer~\cite{he2015delving}. 
In the meantime, we utilize the Front Layer Normalization (Front-LN) in the FedDT-TTS. 
Following the residual block in ResNetV2~\cite{he2016identity} and its symmetric structure explanation, the Front-LN method puts the layer normalization inside the residual connection and equips it with an additional final-layer normalization before prediction. 
This position change of layer normalization would greatly help to converge faster and more steadily. 
Similarly, in the structure of the transformer, we put the layer normalization in front of the Multi-Head Attention and Feed Forward Network (FFN)~\cite{xiong2020layer}.

Our contributions are as follows:(1) We propose the Dynamic Transformer, where encoder and decoder increase layers dynamically. 
This method enables the model to learn the low-level features better, thus greatly improving transformer models' performance in the federated learning, reducing total training time by 40\%. (2) We introduce a simple implementation of weight normalization and the Front-LN approach to achieve a faster and more stable convergence process. 

\begin{figure}[!htb]
  \centering
    \subfigure[The pipeline of a typical TTS system ]{\includegraphics[width=0.43\textwidth]{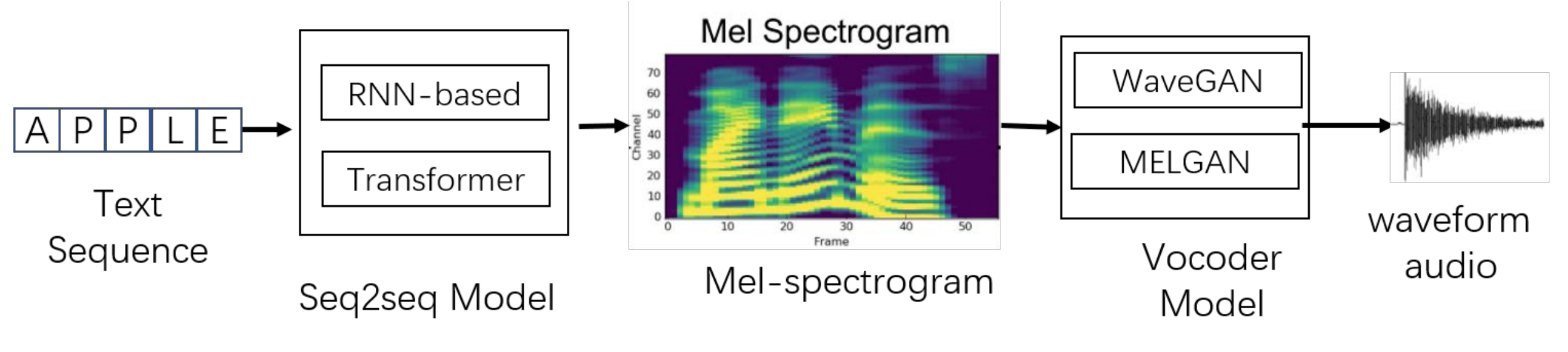}}
    \subfigure[The diagram of our Dynamic Transformer for TTS ]{\includegraphics[width=0.42\textwidth]{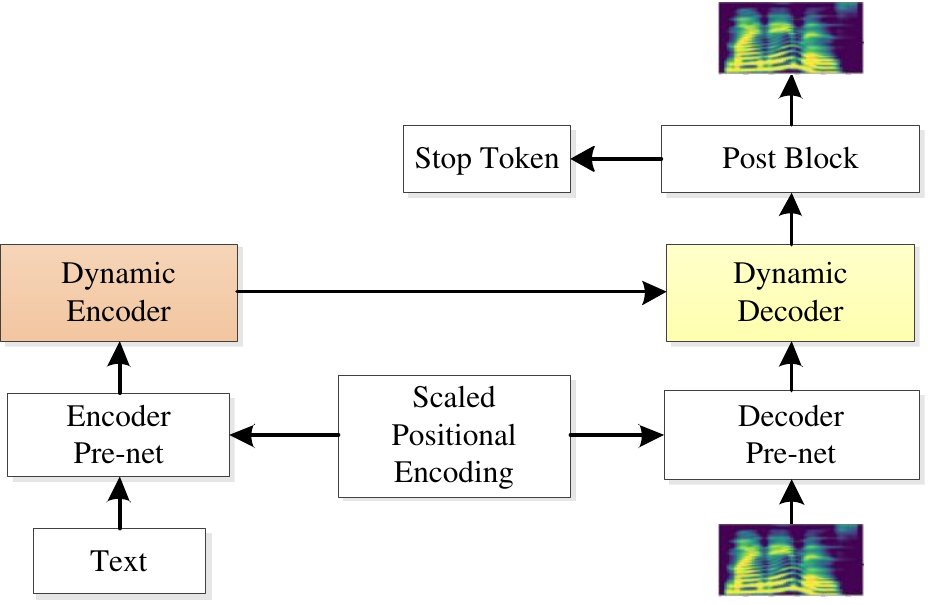}}
    \subfigure[The progressive growing of the encoder/deocder block during communication rounds.]{\includegraphics[width=0.42\textwidth]{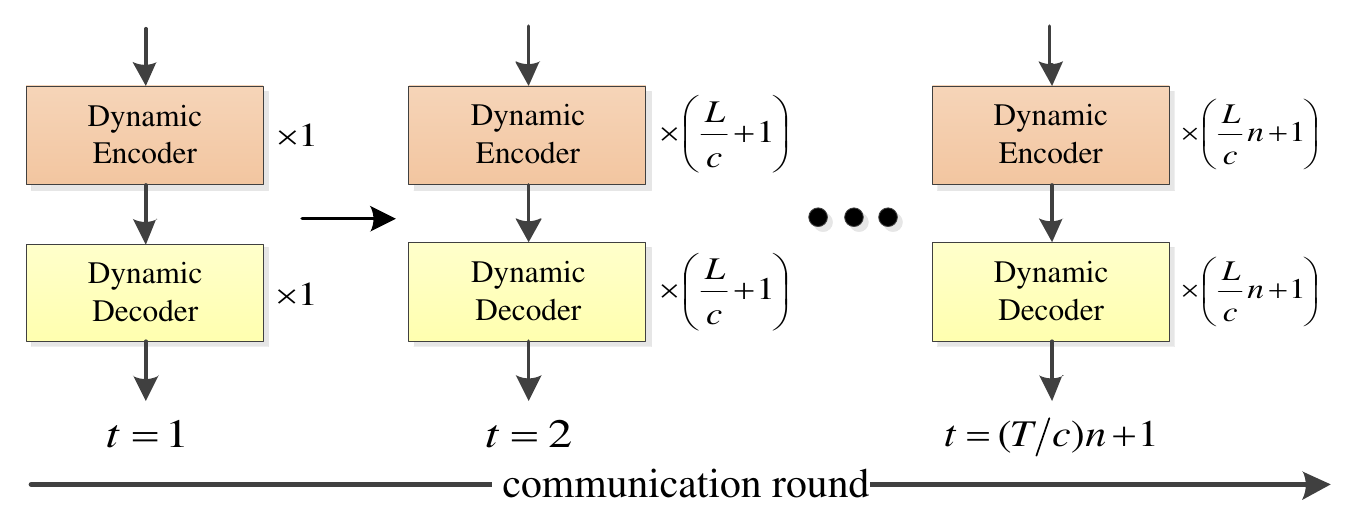}}
    \subfigure[The diagram of our Federated Dynamic Transformer for TTS. Note that the lock icon means encryption and the unlock icon means decryption.]{\includegraphics[width=0.42\textwidth]{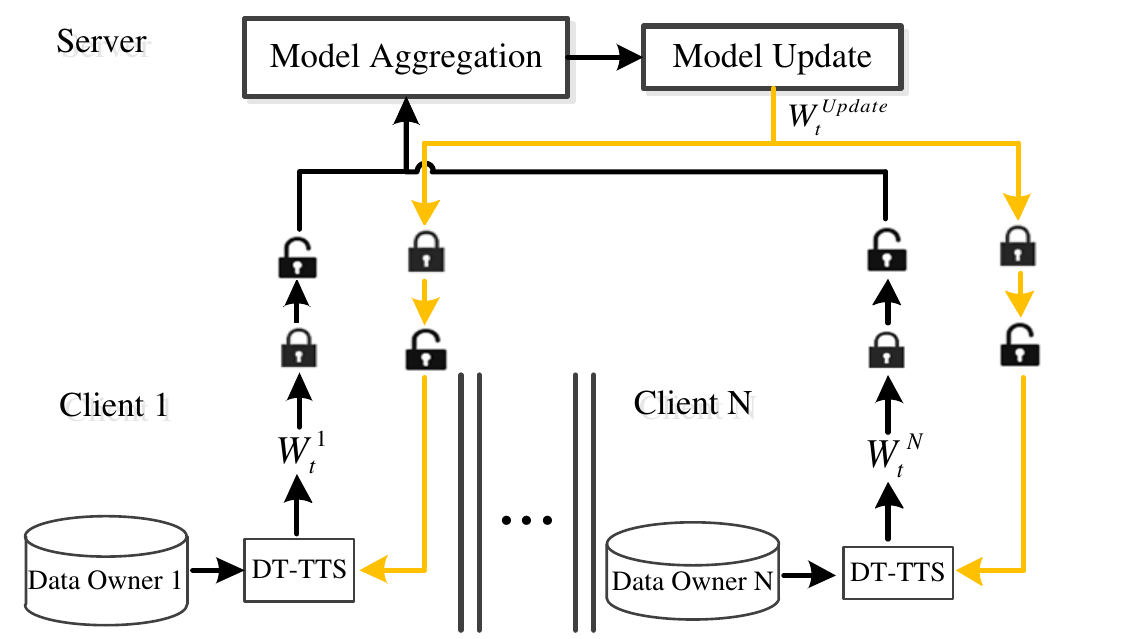}} 
    \vskip -0.15in
    \caption{The diagram of Dynamic Transformer(DT) and Federated Dynamic Transformer(FedDT) for TTS}
    \label{fig:fedtts}
    \vskip -0.1in
\end{figure}

\section{Proposed Method}
\label{sec:proposedmethod}

\subsection{Dynamic Transformer for TTS}
\label{sec::dynamictransformer}

As shown in Figure\ref{fig:fedtts}(a), the typical TTS system comprises two separate modules: a seq2seq model and a vocoder model. 
The pipeline of the TTS system is shown as follows. 
First, the text sequence is fed into the seq2seq model, and the output of the seq2seq model is Mel-spectrograms. 
Then, a vocoder is deployed to convert the spectrogram into time-domain waveforms. 

Here, we use Transformer as the seq2seq model because Transformer TTS enables parallel training and provides an opportunity for building long-range dependencies directly. 
However, as the transformer model's model size hinders the decentralized training and deployment in the mobile devices, we propose  Dynamic Transformer (DT). 
As illustrated in Figure~\ref{fig:fedtts}(b), we used a dynamic encoder and a dynamic decoder. 
The details of Text pre-net are described in~\cite{li2019neural}. 
To be specific, the Text Pre-net contains an embedding layer and an encoder pre-net, generating a vector for the transformer encoder. 
Three fully connected layers compose the Mel Pre-net, converting Mel-spectrograms to vectors as well.

The dynamic growing encoder/decoder blocks process is shown in Figure~\ref{fig:fedtts}(c), where the number $l_{t}$ of layers is following the communication round $t$.
As an example, when $t$ is equal to a number like $2$, then the $l_{2} = (L/c+1)$, where $L$ is the target layers number, $c$ means that L is divided into c parts, and each part has $L/c$. 

In the early stage, the generation of low-level features is substantially more stable because there are less class information and fewer modes. 
By increasing the layers step by step, we are continuously asking for a much higher-level feature. 
Another benefit is the reduced training time. 
With DT, most of the lower-level features are learned at early periods, and considerable performance is often obtained faster in training time. 
The reason is dynamic transformer being in the Taylor-expansion way: the starting terms account most as low ordering ones, while the subsequential ones can refine the function well. 
Hence, adding more layers makes the synthesized wave more natural, since it does better in processing spectrogram details. 

When adding layers, it occurs a sudden shock to the already well-trained shallow layers. 
The new layer with randomly initialized weights would make the loss oscillation. 
So it will let the performance decline to some degree for the first time. 
To stabilize the training process, we make two changes in model structure and weight normalization, respectively. 
Firstly, utilize Front-LN in the transformer architecture. 
We explicitly scale the weights at runtime. 
To be precise, we set $\hat{w} = w/z$, where $w$ are the weights and $z$ is the per-layer normalization constant from He's initializer~\cite{he2015delving}. 
The benefit of doing this dynamically is somehow subtle but relates to the scale-invariance in commonly used adaptive stochastic gradient descent methods such as RMSProp~\cite{tieleman2012divide} and Adam~\cite{DBLP:journals/corr/KingmaB14}. 
These methods normalize a gradient update with its estimated standard deviation, thus making the update independent of the scale of the parameters. 
As a result, if some parameters have a more extensive dynamic range than others, they will take more time to adjust. 
These dues to a scenario modern initializers cause, and thus the learning rate may be both too large and too small at the same time.

\subsection{Federated Dynamic Transformer for TTS}
\label{sec::federateddynamictransformer}

To ensure the security and privacy of original data, we introduce federated learning to the Text-to-Speech synthesis.  
As the communication overhead is the bottleneck of the federated learning, we deploy our Dynamic Transformer for TTS model training in a federated setting, as shown in Figure~\ref{fig:fedtts}(d). 
Following the classical federated learning optimization method FedAvg~\cite{fedavg}, in each communication round, firstly, the server randomly choose $M$ clients composing $S_{t}$, then send the encrypted current global model's weights to every client in $S_{t}$. 
Each client decrypts it and updates the weights using its local data. 
Secondly, clients upload their updated weights to the server through the same encrypt-decrypt procedure. 
After collecting updates, the server would do model aggregation by averaging all clients' updates together, then update the global model with the aggregation value. 

\begin{algorithm}
\caption{Federated Dynamic Transformer}  
\label{alg:A}
\begin{algorithmic}
\STATE {\textbf{Input:} \textit{total communication round T, clients number M, $q$ layers to increase, interval sets of progressive steps $\mathbb{K}$, weights $w$, transformer model f, the target layers number L, client's local iteration steps $I$.}} 
\STATE {\textbf{Server executes:}}
\FOR{each round t = 1,2,..., T}
\STATE{$S_{t} \leftarrow$ (randomly chosen M clients)} \\
$k = t * I$
\IF{$(k \in \mathbb{K})$ and $(l_{t} < L)$}
\STATE{adding layers: $l_{t} \leftarrow l_{t} + q$}
\ENDIF
\FOR{each client $i \in S_{t}$ \textbf{in parallel}}
\STATE{$w^{i}_{t+1} \leftarrow$ \textbf{ClientUpdate}$(i,w_{t}, l_{t})$}
\ENDFOR
\STATE{$w_{t+1} \leftarrow \sum^{M}_{k=1}\frac{1}{M}w^{i}_{t+1}$}
\ENDFOR
\vspace{3mm}
\STATE {\textbf{ClientUpdate$(i,w_{t}, l_{t})$:}}
\IF{adding layers}
\STATE{$f(w,l) \leftarrow f(w,l_{t})$}
\ENDIF
\FOR{local step $j = 1,..,I$}
\STATE{$w \leftarrow w - \eta \nabla f(w,l)$}
\ENDFOR
\STATE{return $w$ to server}
\end{algorithmic}
\end{algorithm}

The federated dynamic transformer(FedDT) is described in Algorithm~\ref{alg:A}.
In clients, if adding layers, renew the $f(w,l)$ with $f(w,l_{t})$. 
In the whole training process, the trained layer's weight is keeping after adding the new layer. 
The benefit of the dynamically growing process when training transformer in the federated learning setting contains two aspects. 

First, both encoder and decoder only have shallow layers at the beginning rounds, which massively reduces the communication cost.
Even the layer number equals the target layer number in the latter communication rounds, the communication cost in the whole training process is much less than training all layers from beginning to end. 
Here is a simple analysis of the communication efficiency of FedDT, following the same setting in Algorithm~\ref{alg:A}.
For FedT, assuming the amount of layers in encoder and decoder is $N$, weights of encoder block and decoder block are $W_{1}$ and $W_{2}$. 
Then the total communication cost of FedT is $T(NW_{1} + NW_{2}) = TN(W_{1} + W_{2})$.
For FedDT, communication cost of each round is $\frac{T}{c}\left[(\frac{N}{c}n+1)W_{1}+(\frac{N}{c}n+1)W_{2}\right]$, where $n$ is from 0 to $c-\frac{c}{N}$. 
Because it's an arithmetic sequence, the sum of all rounds is equal to $\frac{T}{2c}\left[(W_{1}+W_{2})+(NW_{1}+NW_{2})\right] = \frac{T}{2c}(N+1)(W_{1}+W_{2})$.  
Comparing these two total communication cost, FedDT reduces by $\frac{1}{2c}+\frac{1}{2cN}$ ratio theoretically. 
If let $N+1 \approx N$ for simplicity, the reduction ratio is close to $\frac{1}{2c}$.

Second, FedDT's convergence becomes faster and more stable than FedT. 
The reason is that adding layers progressively can make the model's training begin at a good initial weight, thus converging easily. 
Similar to the layer-wise training process in the wake-sleep algorithm, 
when a layer is trained after specific iterations, it gains a well weight distribution. 
Then add a new layer above it. 
The trained layer can be seen as a good initial weight to the new layer. While training the new layer, the old layer can be seen as a  stable and well-trained feature extractor. 
Facing the complexity of the federated learning framework helps the training process become more robust.

\section{Experiments}
\label{sec:experiments}

\subsection{Dataset}
\vskip -0.05in
All experiments are trained on the LJSpeech data\cite{ito2017lj}. 
This dataset consists of 13,100 short audio clips of a single speaker reading passages from 7 non-fiction books.
Clips vary in length from 1 to 10 seconds and have a total length of approximately 24 hours.
The texts are normalized and inserted with the beginning character (a space) and the end character (a period), e.g. ``I am 10 years old'' is converted to `` i am ten years old.''. 
The LJSpeech dataset is randomly divided into two sets: 12600 samples for training and 500 samples for testing.

Separate the training dataset into five subsets, each having 2520 samples. 
Each client owns one subset, which has a length of 4.6 hours roughly. 
Additionally, the test set has a length of almost 1 hour.

\subsection{Settings}
\vskip -0.05in
\textbf{Model Architecture:} 
The transformer model is following~\cite{li2019neural}. 
The DT-TTS and Transformer-TTS configurations are as follows: the hidden size, attention head, FFN filter size are 384, 4, 1536 respectively. 
These hyper-parameters are also the same as when fitted in federated learning.  
For the Transformer-TTS, the layers of encoder and decoder are fixed to 6.
The inputs for models were all texts and Mel-spectrograms.

\noindent\textbf{Compare Schemes:} 
We analyze our model in two schemes: (i) A standalone case comparing Transformer-TTS and our DT-TTS, both are trained on the entire trainset in a single worker; (ii) Considering Transformer-TTS in (i) as a baseline, we evaluate both Transformer-TTS and our DT under the federated learning framework. 

\noindent\textbf{Evaluation:} 
To quantify models' performances, we use the mean opinion score (MOS) to measure generated voices' qualities. 
Twenty native English speakers judge each sentence.
There are ten males and ten females.

\noindent\textbf{Training Strategy:} 
We experimented with the standalone case with 4 NVIDIA V100 GPUs and trained the FedDT-TTS with 1 NVIDIA V100 GPU on each client. 
The total communication rounds $T$ is 120, and target layers $L$ for both encoder and decoder are assigned to 6.
We set the $c$ equal to 6. 
Then the $L/c$ is 1, and the $T/c$ is 20, that means we add 1 layer for both the encoder and decoder every 20 communication rounds.
ALL training processes utilize a mini-batch Adam optimizer, setting the batch size to 16 for every client.
In addition, RSA encryption is used to protect clients' privacy.

\subsection{Analysis}
\vskip -0.05in
\textbf{The Standalone Case:} In Table~\ref{tab:table1}, both models are trained on the entire trainset. 
Our DT-TTS converages almost $20\%$ faster than the T-TTS, which trains the transformer model with six layers from beginning to end.
From the perspective of MOS, DT-TTS is $4.03\pm0.11$, which is nearly close to T-TTS ($4.01\pm0.07$).
Our model achieves a comparable performance and reduces the training time appreciably.
\begin{table}[h]
\vskip -0.1in
	\centering
	\caption{Comparison of Average Convergence Steps and MOS.}
	\label{tab:table1}
	\begin{tabular}{ccc}
		\toprule
		\textbf{Model} & \textbf{\begin{tabular}[c]{@{}l@{}}Average Convergence Steps \end{tabular}}
		& \textbf{\begin{tabular}[c]{@{}l@{}}MOS \end{tabular}}\\ 
		\midrule
		T-TTS & 697125 & 4.03$\pm$0.11 \\
		DT-TTS & 539833 & 4.01$\pm$0.07 \\
		\bottomrule
	\end{tabular}
	\vskip -0.05in
\end{table}

\begin{figure}[t]
\vskip -0.2in
	\centering
	\includegraphics[width=0.9\linewidth]{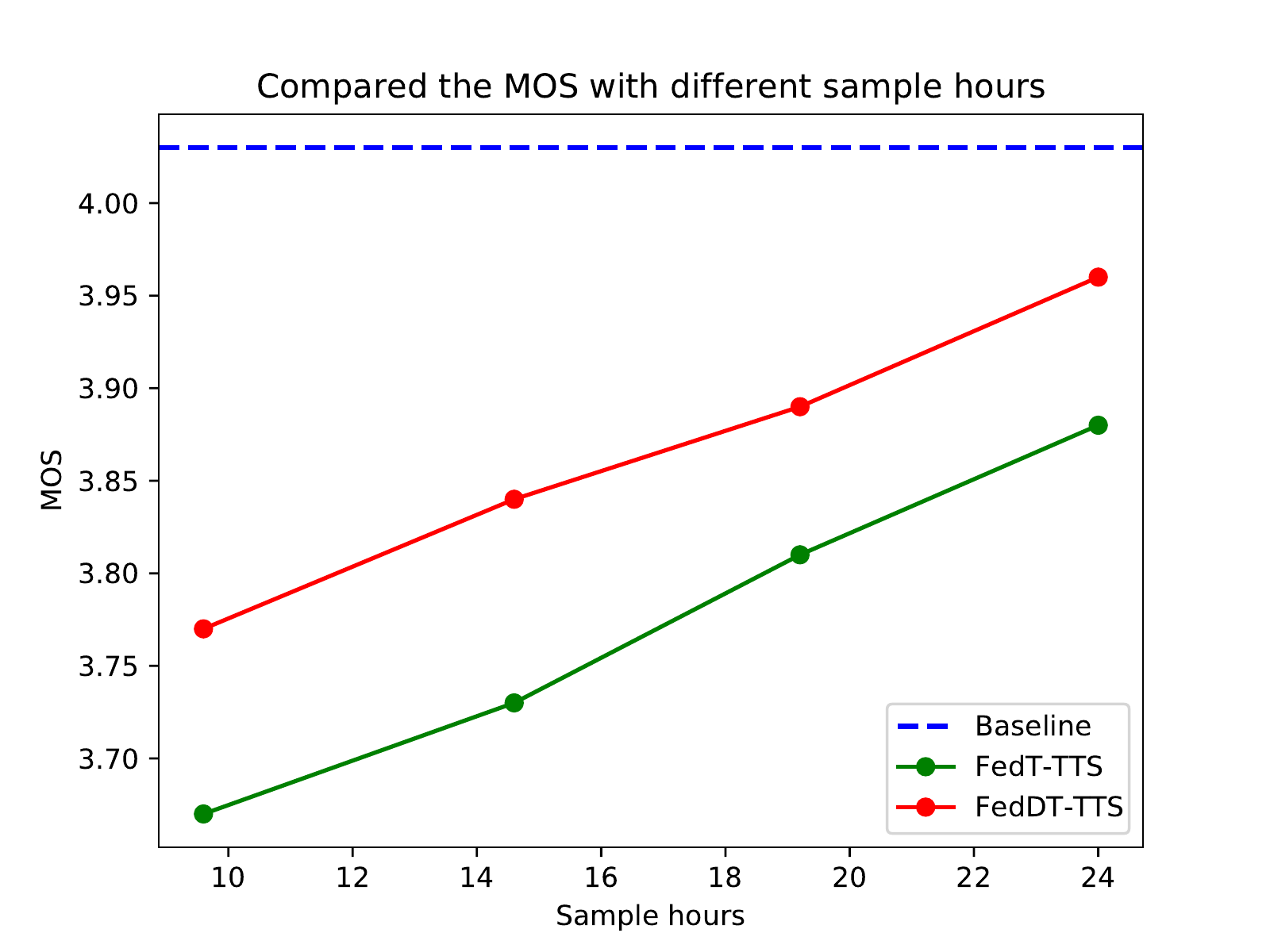}  
	\caption{Models' comparison when increasing sample datas.} 
	\label{fig:sampeltime}
	\vskip -0.2in
\end{figure}

\noindent\textbf{In Federated Learning:}
As a visual illustration in Figure~\ref{fig:sampeltime}, FedT-TTS and FedDT-TTS both improve their performances when the number of clients increases. 
Obviously, federated learning greatly helps by collaborating more data from other parties.
In Table~\ref{tab:table2}, we detail the MOS comparison between FedT-TTS and FedDT-TTS.
We can see that when increasing the client's number, MOS for both models is ascending, our FedDT-TTS is approaching baseline's performance when all clients participating (3.96 to 4.03), especially. 
Besides, ours outperforms FedT-TTS in terms of client's numbers.
On the other hand, ours hugely reduces the total training time by around $20\%$ in different clients number with the superior performance. 
Especially in a 5-clients situation, FedDT-TTS's time cost is 49.26h, almost $30\%$ reduction compared with FedT-TTS, 66.42h.
We can see an almost linear reduction of FedDT-TTS according to the increasing number of clients. 
This majorly owes to the fewer parameters brought by the dynamic process, reducing training time and communication time. 
Above all, these significantly demonstrate that ours is more stably trained and time-efficient than the FedT-TTS.

\begin{table}[htbp]
\centering
\caption{Time cost and MOS comparision between FedT-TTS and FedDT-TTS. ``M'' is the number of clients.}
\label{tab:table2}
\vskip -0.05in
\begin{tabular}{ccccc}
\toprule
\textbf{Model} & \textbf{\begin{tabular}[c]{@{}l@{}}M \end{tabular}}
& \textbf{\begin{tabular}[c]{@{}l@{}}Time Cost \end{tabular}}
& \textbf{\begin{tabular}[c]{@{}l@{}}MOS \end{tabular}}\\ 
\midrule
T-TTS & - & - & 4.03$\pm$0.11 \\ 
\midrule
FedT-TTS & 2 & 23.54h & 3.67$\pm$0.11 \\
FedDT-TTS & 2 & 18.35h & \textbf{3.77$\pm$0.11} \\ 
\midrule
FedT-TTS & 3 & 33.86h & 3.73$\pm$0.10 \\
FedDT-TTS & 3 & 26.44h & \textbf{3.84$\pm$0.05} \\ 
\midrule
FedT-TTS & 4 & 48.16h & 3.81$\pm$0.08 \\
FedDT-TTS & 4 & 37.47h & \textbf{3.89$\pm$0.07} \\ 
\midrule
FedT-TTS & 5 & 66.42h &  3.88$\pm$0.06 \\
FedDT-TTS & 5 & 49.26h & \textbf{3.96$\pm$0.04} \\
\toprule
\end{tabular}
\vskip -0.15in
\end{table}


\noindent\textbf{On Unbalanced Dataset:}
We also compare FedT-TTS and FedDT-TTS on an unbalanced dataset, which commonly happens in real federated learning applications.
We investigate in a 3-clients setting, resplit the whole trainset into three subsets according to the split ratios, then assign each subset to each client. 
It shows that on the unbalanced dataset, ours still outperforms the FedT-TTS.
The average improvement is about +0.09 for all split ratios.  
These results suggest that the FedDT-TTS is more suitable in federated learning, even on the unbalanced dataset.

\begin{figure}[t]
	\centering
	\includegraphics[width=0.9\linewidth]{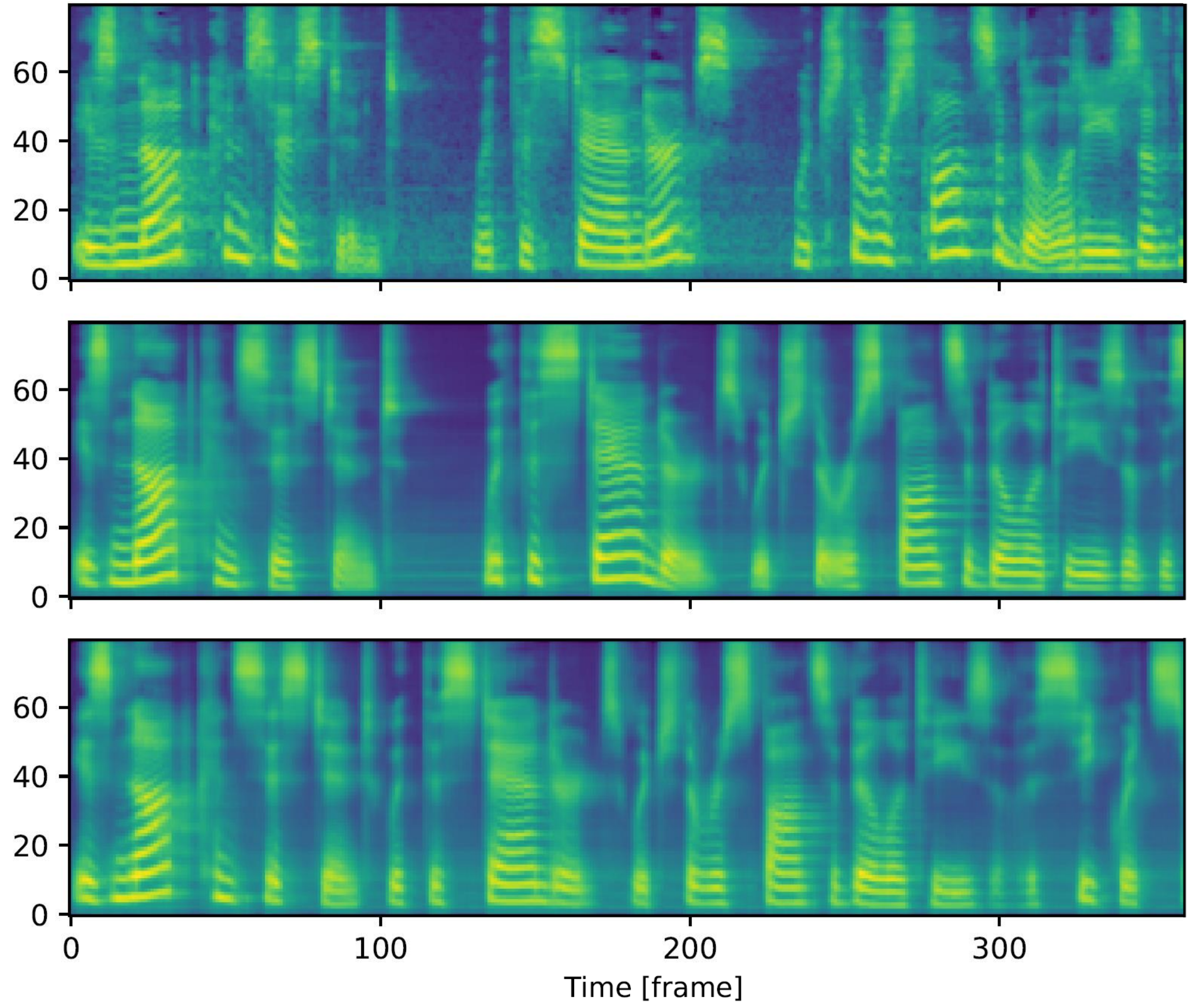}
	\caption{Visual comparison of mel-spectrograms. (top) ground-truth, (middle) T-TTS, (bottom) DT-TTS.} 
	\label{fig:melspectrogram}
	\vskip -0.15in
\end{figure}

\begin{table}
\centering
\caption{MOS comparision between FedT-TTS and FedDT-TTS
``r'' is the ratio of spliting with $M$ being 3.}
\label{table3}
\begin{tabular}{cccc}
\toprule
\textbf{Model} & \textbf{\begin{tabular}[c]{@{}l@{}}$r$ of 3 clients \end{tabular}}
& \textbf{\begin{tabular}[c]{@{}l@{}}MOS \end{tabular}}\\ \midrule
FedT-TTS & 1:1:1 & 3.73$\pm$0.10 \\
FedDT-TTS & 1:1:1 & \textbf{3.84$\pm$0.05} \\ 
\midrule
FedT-TTS & 1:1:3 & 3.66$\pm$0.11 \\
FedDT-TTS & 1:1:3 & \textbf{3.74$\pm$0.09} \\ 
\midrule
FedT-TTS & 1:2:4 & 3.68$\pm$0.10 \\
FedDT-TTS & 1:2:4 & \textbf{3.78$\pm$0.05} \\
\midrule
FedT-TTS & 1:1:5 & 3.63$\pm$0.08 \\
FedDT-TTS & 1:1:5 & \textbf{3.72$\pm$0.06} \\
\midrule
FedT-TTS & 1:1:6 & 3.66$\pm$0.07 \\
FedDT-TTS & 1:1:6 & \textbf{3.75$\pm$0.04} \\
\midrule
FedT-TTS & 1:1:8 & 3.70$\pm$0.07 \\
FedDT-TTS & 1:1:8 & \textbf{3.79$\pm$0.05} \\
\toprule
\end{tabular}
\vskip -0.25in
\end{table}

\noindent\textbf{Mel-Spectrograms Visualization:} 
We also analyze Mel-spectrograms generated by DT-TTS and T-TTS, respectively, inputting the same text, and compare them together with ground truth, as shown in Figure~\ref{fig:melspectrogram}. 

\section{Conclusion}
\vskip -0.05in
\label{sec:conclusion}
In this paper, we propose the Dynamic Transformer for TTS synthesis.
Comparing with T-TTS, ours shows a significant reduction in training time with miniature MOS deducted.
Combining with the federated learning framework, we improve the model's performance by collaborating with more data owners under absolute data privacy protection. 
The MOS of our model keeps increasing while adding clients, approaching the baseline method gradually. 
Besides, experiments on the LJSpeech datasets show that our FedDT-TTS surpasses FedT-TTS in time efficiency and model performance.  
On the one hand, the dynamic adding layers process significantly saves client's training time and communication costs. 
On the other hand, it contributes to improving global convergence robustly.  
Moreover, experiments on an unbalanced dataset and Mel-spectrograms analysis also prove that our DT-TTS is superior to Transformer-TTS in the federated learning framework. 

\section{Acknowledgement}
\vskip -0.05in
This work is supported by National Key Research and Development Program of China under grant No.2018YFB0204403, No.2017YFB1401202 and No.2018YFB1003500. Corresponding author is Jianzong Wang from Ping An Technology (Shenzhen) Co., Ltd.

\bibliographystyle{IEEEtran}
\bibliography{interspeech2021-fedtts}

\end{document}